\pdfoutput=1
\PassOptionsToPackage{table,dvipsnames}{xcolor}
\documentclass[11pt]{article}

\usepackage[final]{latex/acl}

\usepackage{times}
\usepackage{latexsym}
\usepackage[table]{xcolor}
\usepackage[dvipsnames]{xcolor}
\usepackage{makecell}
\usepackage[T1]{fontenc}

\usepackage[utf8]{inputenc}

\usepackage{microtype}

\usepackage{inconsolata}

\usepackage{inconsolata}
\usepackage{latexsym}
\usepackage{float}
\usepackage{mathrsfs}
\usepackage{amsthm,amsmath,amssymb}
\usepackage{multirow}
\usepackage[ruled]{algorithm2e}
\usepackage{threeparttable}
\usepackage{array}
\usepackage{booktabs}
\usepackage{bm}
\usepackage{graphicx}
\usepackage{amssymb}
\usepackage{wasysym}
\usepackage{bbm}

\definecolor{Mycolor1}{HTML}{FDC3D2}
\definecolor{Mycolor2}{HTML}{DCFDC4}
\title{Expanding before Inferring: Enhancing Factuality in Large Language Models through Premature Layers Interpolation}

\author{
Dingwei Chen{$^{1,2}$}, Ziqiang Liu{$^{5}$}, Feiteng Fang{$^{5}$}, Chak Tou Leong{$^{3}$}, Shiwen Ni{$^{5}$}, \\  \textbf{Ahmadreza Argha}{$^4$}, \textbf{Hamid Alinejad-Rokny}{$^4$}, \textbf{Min Yang}$^{5*}$, \textbf{Chengming Li}{$^{2}\thanks{~~Corresponding author.}$} \\
\small $^1$ Sun Yat-Sen University $^2$Shenzhen MSU-BIT University
$^3$The Hong Kong Polytechnic University \\
\small $^4$UNSW, Sydney, NSW 2052, Australia 
\small $^5$Shenzhen Key Laboratory for High Performance Data Mining, \\ \small Shenzhen Institutes of Advanced Technology, Chinese Academy of Sciences 
\\ \small \texttt{cuso4cdw@gmail.com, licm@smbu.edu.cn, min.yang@siat.ac.cn}
  }
  
\begin{document}
\maketitle
\begin{abstract}
Large Language Models (LLMs) demonstrate remarkable capabilities in text understanding and generation. However, their tendency to produce factually inconsistent outputs—commonly referred to as ``hallucinations''—remains a critical challenge. Existing approaches, such as retrieval-based and inference-time correction methods, primarily address this issue at the input or output level, often overlooking the intrinsic information refinement process and the role of premature layers. Meanwhile, alignment- and fine-tuning-based methods are resource-intensive.
In this paper, we propose \textbf{PLI} (\textbf{\underline{P}}remature \textbf{\underline{L}}ayers \textbf{\underline{I}}nterpolation), a novel, training-free, and plug-and-play intervention designed to enhance factuality. PLI mitigates hallucinations by inserting premature layers formed through mathematical interpolation with adjacent layers. Inspired by stable diffusion and sampling steps, PLI extends the depth of information processing and transmission in LLMs, improving factual coherence.
Experiments on four publicly available datasets demonstrate that PLI effectively reduces hallucinations while outperforming existing baselines in most cases. Further analysis suggests that the success of layer interpolation is closely linked to LLMs' internal mechanisms. Our dataset and code are available at \href{https://github.com/CuSO4-Chen/PLI}{\texttt{https://github.com/CuSO4-Chen/PLI}}.
\end{abstract}

\section{Introduction}
Recently, Large Language Models (LLMs) have revolutionized artificial intelligence with their unprecedented capabilities in language understanding and open-ended generation, demonstrating strong performance across various downstream tasks \cite{guo2025deepseek, brown2020language, achiam2023gpt, zhao2023survey}. However, their remarkable progress is overshadowed by a persistent challenge: the tendency to generate plausible yet factually incorrect content, commonly referred to as ``hallucinations'' \cite{zhang2024truthx, chuang2023dola}. These deficiencies in factual grounding undermine the reliability of LLM applications, making hallucination mitigation a critical area of research \cite{huang2023survey, yang2024improving, kai2024sh2, chen2024lower}. Prior studies suggest that hallucinations stem from multiple factors, including low-quality large-scale pretraining data \cite{zhang2023siren, ye2023cognitive}, model training errors \cite{zhang2023alleviating}, and unstable decoding strategies during generation \cite{huang2023survey, cheng2025think}. 

\begin{figure}[t]
	\centering
\includegraphics[width=1\linewidth]{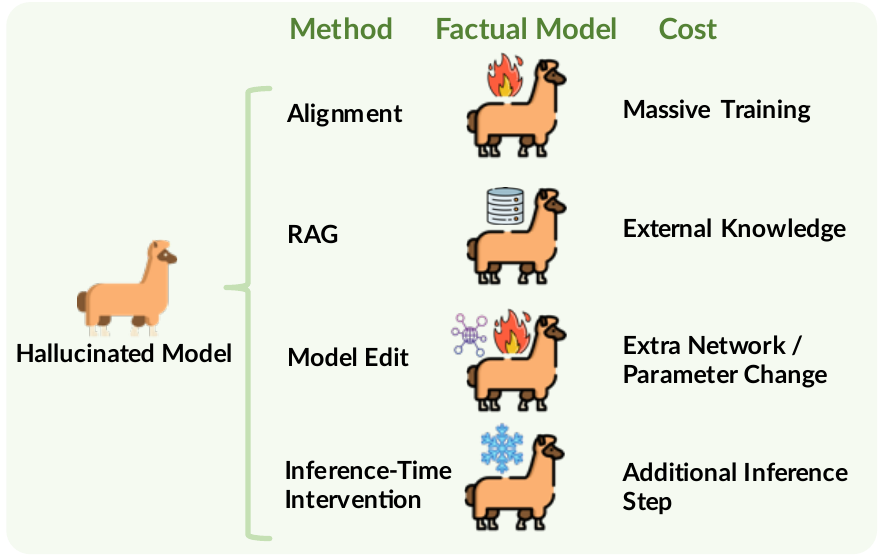}
	\caption{Brief overview of previous methods for alleviating hallucination in LLMs.
	}
	\label{f1}
\end{figure}

Existing approaches (in Figure \ref{f1}) to mitigating hallucinations typically leverage external knowledge bases \cite{peng2023check, jiang2023active, wang2023knowledgpt}, align models with human feedback \cite{ouyang2022training}, refine output distributions through contrastive decoding (CD) \cite{li2023contrastive, chuang2023dola, zhang2023alleviating, o2023contrastive, chen2024lower}, modify internal knowledge or output representations \cite{meng2023mass, ni2023forgetting, zhang2024truthx, liang2024locate}. Among these, inference-time methods are gaining increasing attention due to their simplicity and lower computational cost, but still with extra inference steps.

Empirically, LLMs process information through successive transformer layers, where each layer refines the previous representation toward greater semantic and factual coherence \cite{lad2024remarkable}. Thus, as information flows from input to output, it can be viewed as progressively processing and refining knowledge. Regarding how existing approaches mitigate hallucination, retrieval-augmented generation (RAG) \cite{lewis2020retrieval} enhances the input stage by incorporating external knowledge, while inference-time methods (e.g., CD) refine the output. However, little research has focused on optimizing the intermediate representations of \textit{premature layers} (i.e., intermediate layers), which could help mitigate hallucinations in the final output. While injecting additional modules, such as adapters \cite{houlsby2019parameter} or transformer blocks \cite{wu2024llama}, may achieve similar effects, these approaches require extra training costs. Efficiently optimizing the output representations of premature layers with minimal overhead remains an open challenge.

To this end, we propose a novel training-free paradigm for hallucination alleviation called \textbf{PLI} (\textbf{\underline{P}}remature \textbf{\underline{L}}ayers \textbf{\underline{I}}nterpolation).
Our method stems from the observation that LLMs typically process and refine knowledge in a fixed number of steps (i.e., the number of layers). This contrasts with generative models like diffusion models \cite{rombach2022high}, which allow for adjustable refinement steps where more steps generally lead to higher quality generation. PLI aims to extend the knowledge processing depth in LLMs to reduce hallucinations by effectively increasing the number of model layers.
Specifically, PLI is a simple yet effective method that strategically inserts one or more parameter-interpolated layers. These layers are constructed by interpolating the parameters of two adjacent layers, expanding the refinement process, and enhancing the inherent capacity of LLMs for factual consistency.
Furthermore, PLI is designed as a plug-and-play module that can seamlessly integrate with existing base models and hallucination mitigation techniques. The main contributions of this work can be summarized as follows:

\begin{itemize}
\item We propose PLI, a novel framework for hallucination alleviation that is both training-free and plug-and-play. PLI leverages the existing parameter manifold through interpolation to construct new premature layers, enhancing factuality in LLMs.
\item We conduct extensive experiments on four widely used benchmarks, demonstrating that PLI outperforms baseline methods in most cases and can be effectively integrated with existing hallucination mitigation techniques.
\item We perform a series of analyses to theoretically investigate the relationship between layer interpolation effectiveness and the internal mechanisms of LLMs.
\end{itemize}


\section{Related Work}
\subsection{Halluciantion Mitigation}
Current approaches to mitigate hallucinations can be broadly categorized into several classes~\cite{huang2023survey, zhang2023siren}.

\paragraph{Alignment with Feedback} plays a crucial role in training LLMs, ensuring that models adapt to specific feedback and effectively reduce hallucinations \cite{liu2023chain, menick2022teaching}.
\citet{stiennon2020learning} construct a high-quality corpus with human feedback to train models that generate human-preferred summaries, promoting more factual and reliable outputs. 
\citet{shinn2023reflexion} propose a method using verbal reinforcement to help agents learn from failures through self-reflection.

\paragraph{Retrieval-Augmented Generation (RAG)} is a powerful approach that enhances LLMs by integrating factual knowledge, thereby improving factual accuracy~\cite{wang2023knowledgpt, li2023large, liu2023reta}. \citet{peng2023check} introduce a framework that augments LLMs with external knowledge and automated feedback, helping to generate more truthful responses. \citet{trivedi2023interleaving} propose a retrieval method that leverages LLMs’ chain-of-thought (CoT) reasoning capabilities to reduce hallucinations and enhance logical consistency. 

\paragraph{Model Editing} refers to modifying the knowledge stored in LLMs, enabling data-efficient updates to correct hallucinations or outdated information~\cite{zheng2023can, huang2023transformer, gupta2023editing, yao-etal-2023-editing, ni2023forgetting, zhang2024comprehensivestudyknowledgeediting}. SERAC~\cite{mitchell2022memory} routes new facts through a distinct network while keeping the original parameters unchanged. IKE~\cite{zheng2023can} edits the model by prompting it with the revised fact using in-context examples. 
ROME~\cite{meng2022locating} modifies specific knowledge neurons in feed-forward networks (FFNs) through a locate-then-edit approach 
MEMIT\cite{meng2023mass} extends this by simultaneously updating multiple pieces of knowledge. 

\subsection{Inference-Time Intervention}
\textbf{Representation editing} is an inference-time method that modifies a model's internal representations to improve output factuality~\cite{li2023inference, zhang2024truthx, chen2024truth, burns2022discovering}. ITI~\cite{li2023inference} probes and adjusts truthfulness within the attention heads of LLMs. TruthForest~\cite{chen2024truth} employs orthogonal probes to uncover hidden truth representations. TruthX~\cite{zhang2023alleviating} decouples LLM representations into separate truthful and semantic latent spaces using an autoencoder and applies contrastive learning to refine factual outputs.

\textbf{Contrastive Decoding} (CD) is a decoding-time framework that enhances factuality by contrasting predictions with the logits of a much smaller LLM. 
DoLa~\cite{chuang2023dola} applies CD between later and earlier layers to boost factuality and reasoning. ICD~\cite{zhang2023alleviating} penalizes hallucination-prone predictions by inducing contrast with a factually weaker LLM. 
Beyond CD, HaluSearch~\cite{cheng2025think} integrates Monte Carlo Tree Search (MCTS) to enable a deliberate, slow-thinking generation process for mitigating hallucinations 

In this paper, we propose PLI, a novel inference-time method that extends LLMs' end-to-end information processing by inserting premature layers constructed through interpolation. These interpolated layers optimize intermediate representations to reduce hallucinations. PLI is highly flexible and can seamlessly integrate with other hallucination alleviation techniques.

\begin{figure*}[t]
	\centering
\includegraphics[width=0.83\linewidth]{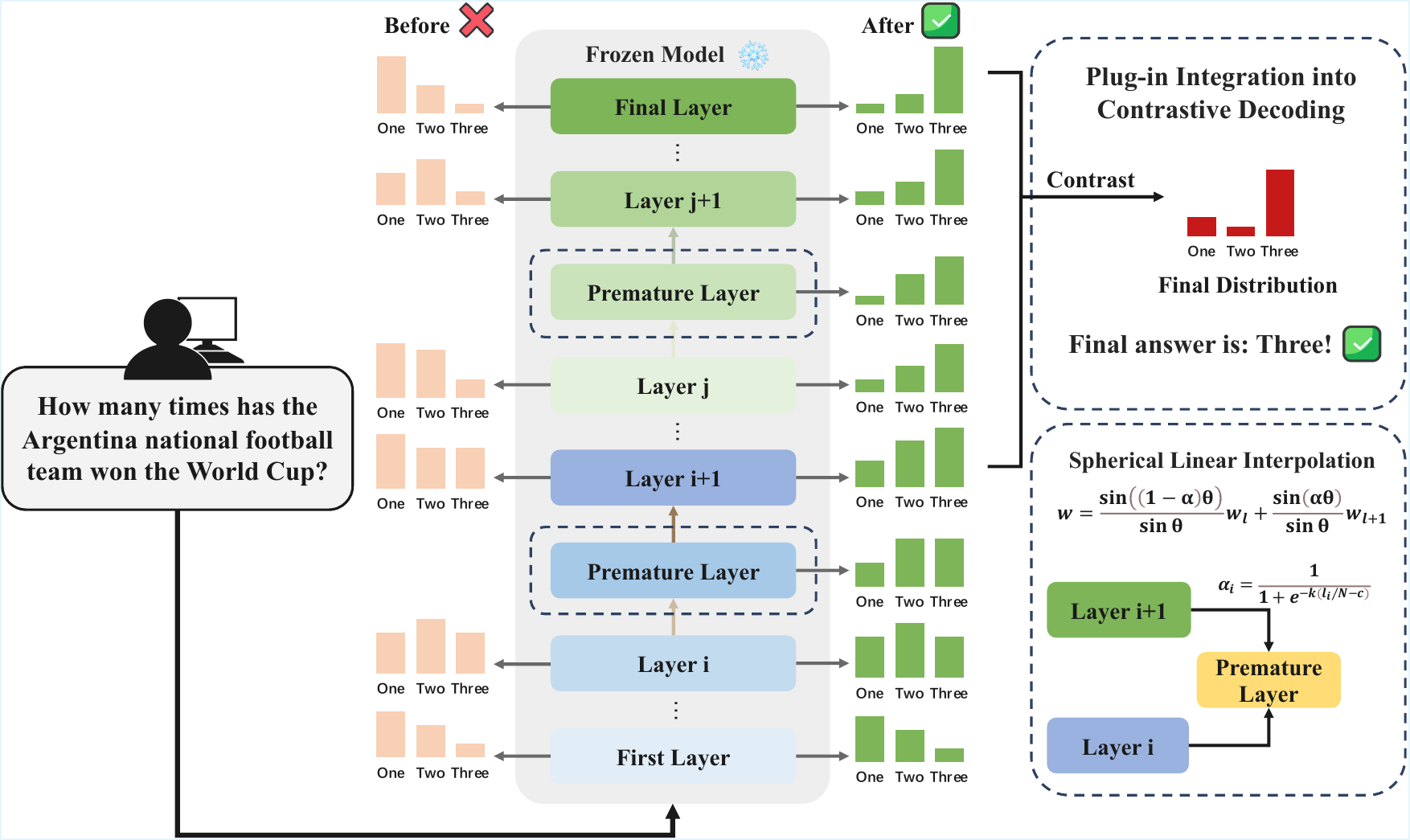}
	\caption{Overview of the \textbf{PLI} method, which inserts premature layers formed by mathematical interpolation to expand the information processing flow, enhancing the factuality and alleviating hallucination in LLMs.}
	\label{f2}
\end{figure*}

\section{Method} \label{Method}
The proposed PLI (Premature Layers Interpolation) method mitigates  hallucination in LLMs by strategically inserting interpolated premature layers between the existing transformer layers. We treat the depth of layers \textit{L} as a controllable dimension, rather than a fixed architectural constant, to enhance factual coherence. PLI uses \textbf{spherical linear interpolation (Slerp)} \cite{robeson1997spherical} to preserve the geometric properties of parameter vectors in high-dimensional space. This ensures smoother transitions between layer representations and optimizes the output representation of the premature layers. Figure~\ref{f2} illustrates the diagram of PLI.

\subsection{Slerp} \label{3.1}
Slerp \cite{robeson1997spherical} is a well-established method for smoothly interpolating between two vectors, commonly applied in fields such as statistical forecasting and model merging~\cite{yang2024model, lu2024merge}. It is calculated as follows:
\begin{equation}
Slerp(p,q,t)=\frac{\sin[(1-t)\theta]\cdot p+\sin(t\theta)\cdot q}{\sin\theta},
\end{equation}
where $p$, $q$ are the vectors to be interpolated,  $\theta$ is the angle between these two vectors, $t$ defines the interpolation ratio. The  division by $sin\theta$ ensures unit normalization.

In a similar manner, we apply Slerp to construct the interpolated premature layers. Given an LLM $f_{\theta}$ with two adjacent layers $l$ and $l+1$, where $l$ is the layer where the position is inserted, we  treat their flattened parameter matrices $W_{l}$ and $W_{l+1}$ as normalized vectors $\mathbf{w}_{l}$, $\mathbf{w}_{l+1}$ $\in\mathbb{R}^d$ on a $d$-dimensional hypersphere. The interpolated layer parameters $\mathbf{w}_{inter}$ are computed as follows:
\begin{equation}
\mathbf{w}_{\mathrm{inter}}=\frac{\sin\left((1-\alpha)\theta\right)}{\sin\theta}\mathbf{w}_l+\frac{\sin(\alpha\theta)}{\sin\theta}\mathbf{w}_{l+1},
\end{equation}
where $\theta$ is the angle between the vectors $w_{l}$ and $w_{l+1}$, and $\alpha$ $\in\mathbb[0, 1]$ is the interpolation ratio. The angle $\theta$ is computed as:
\begin{equation}
\theta=\arccos\left(\mathbf{w}_l\cdot\mathbf{w}_{l+1}\right).
\end{equation}
This interpolation formulation ensures that the interpolated vector $\mathbf{w}_{\mathrm{inter}}$ lies on the same hypersphere as the original vectors, preserving the magnitude and directionality. This property is essential for maintaining the model’s intrinsic knowledge manifold, ensuring smoother transitions between layers in the model.

\subsection{Premature Layers Interpolation Insertion}

For an $N$-layers LLM $f_{\theta} = \{l_1,l_2,...,l_N\}$, we insert interpolated layers $M$ at predefined layer positions, resulting in the modified model $f_{\theta} = \{l_1,l_2,...,l_M,...,l_{N+M}\}$, achieved through premature layer interpolation. The insertion process follows three steps:
\textbf{(i) Adjacent layer pair selection:} For each target insertion position $l_i$, identify the flanking layers ($l_i$, $l_{i+1}$), which refers to the new premature layer being inserted between these two layers.
\textbf{(ii) Slerp Execution:} Compute $\mathbf{w}_{\mathrm{inter}}^{i}$ using Eq. (2) with layer-specific $\alpha_{i}$ to control the interpolation ratio.
\textbf{(iii) Parameter reshaping:} Restructure $\mathbf{w}_{\mathrm{inter}}^{i}$ from the previous step into the original matrix dimensions of the parameters to form the complete parameters for the premature layer interpolation.

After the insertion of premature layers through interpolation, the forward propagation during inference within the LLM will be updated as follows:
\begin{equation}
h^{\prime}=\mathrm{Layer}_{\mathrm{inter}}^{i}(\mathrm{Layer}_{l_i}(h))
\end{equation}
where $h$ is the hidden state from the layer $l_{i-1}$, and $\mathrm{Layer}_{\mathrm{inter}}^{i}$ denotes the inserted premature layer of interpolation. When the hidden state is output from layer $l_i$, it is input to the newly inserted premature layer for further information processing, which is then passed on to layer $l_{i+1}$. This execution can be repeated to insert additional premature layers for more precise hallucination alleviation. Premature layers interpolation is plug-and-play, meaning it can enhance the original model without causing any conflict with the base model or other methods.

\subsection{Adaptive Interpolation Ratio Scheduling}
\label{3.3}

In LLMs, different levels of transformer layers correspond to varying degrees of semantic abstraction. The lower layers (closer to the input end) primarily process local grammatical patterns and vocabulary-level information, while the higher layers (closer to the output end) focus on global semantic integration and factual information \cite{chuang2023dola}. To accommodate this hierarchical semantic distribution, we propose an adaptive interpolation ratio scheduling mechanism for determining $\alpha_{i}$. Specifically, the interpolation ratio $\alpha_{i}$ of an inserted layer is dynamically adjusted based on the position of the inserted layer $l_i$ rather than being fixed. We utilize a variant of the sigmoid function to implement this scheduling:
\begin{equation}
\alpha_i=\frac{1}{1+e^{-k(l_i/N-c)}},
\end{equation}
where $l_i$ $\in\mathbb[0, N-1]$ represents the inserted layer index, and $N$ is the total number of layers within the LLM. The constant $k$ controls the steepness of the sigmoid curve, while $c$ determines the center offset of the function.

When the interpolation layer is inserted in the lower half of the model (i.e., $l_i/N < 0.5$), $\alpha_{i}$ decreases, bringing the interpolation result closer to the parameters of the lower layers, thereby preserving local grammatical and lexical information. As the layer is inserted near the middle (i.e., $l_i/N \approx 0.5$), $\alpha_{i}$ approaches $0.5$, which results in a balanced combination of parameters from both lower and higher layers, merging local and global information. When the interpolation is inserted into the upper layers (i.e. $l_i/N > 0.5$), the interpolation approaches the parameters of the higher layers, emphasizing global semantics and factual consistency. Overall, this scheduling mechanism facilitates smooth transitions among the premature layers and optimizes the consistent flow of information, enhancing both local details and global coherence across the model.

\section{Experiment}

\subsection{Datasets}
We conduct experiments on four widely recognized benchmark datasets:  \textbf{TruthfulQA} \cite{lin2022truthfulqa}, \textbf{FACTOR} \cite{muhlgay2024generating}, \textbf{StrategyQA} \cite{geva2021strategyqa}, and \textbf{GSM8K} \cite{cobbe2021gsm8k}. These datasets are used to assess hallucination mitigation and reasoning capabilities in LLMs.

\textbf{TruthfulQA} is a comprehensive dataset designed to evaluate the factual accuracy of large language models. Following previous work, we adopt a multiple-choice question format for our experiments. The dataset consists of 817 multiple-choice questions spanning 38 diverse domains. Performance is measured using three key metrics: \textbf{MC1}, \textbf{MC2}, \textbf{MC3}, which evaluate the probability of correct and incorrect answers from three different perspectives. \textbf{FACTOR} dataset consists of three sub-datasets—\textbf{News}, \textbf{Wiki}, \textbf{Expert}—which serve as a benchmark for content completion and factuality evaluation. The primary evaluation metric is accuracy, which measures the factual correctness of text completions generated by LLMs.

In addition, to evaluate the effectiveness of our method on generation tasks, we adopt two datasets focused on chain-of-thought (CoT) reasoning: \textbf{StrategyQA} and \textbf{GSM8K}. StrategyQA consists of 2288 question-answer pairs designed to test common sense reasoning and logical inference in language models. GSM8K is a widely used benchmark in the LLM era, containing over 8,000 high-quality, graduate-school-level math problems. It serves as a key standard for evaluating mathematical reasoning and generation capabilities in LLMs. For both datasets, we use accuracy as the primary evaluation metric. Notably, these datasets primarily consist of general knowledge tasks, making them valuable for assessing the overall reasoning abilities of LLMs beyond just hallucination mitigation.

\begin{table*}[t] \small
\setlength\tabcolsep{11pt}
\renewcommand\arraystretch{1.2}
\begin{center}
\begin{tabular}{lcccccccc}

\toprule
\multirow{2}{*}{\textbf{Method}}&\multicolumn{3}{c}{\textbf{TruthfulQA}}&\multicolumn{3}{c}{\textbf{FACTOR}} &\multicolumn{2}{c}{\textbf{CoT}} \\
\cmidrule(r){2-4}  \cmidrule(r){5-7} \cmidrule(r){8-9}
& \textbf{MC1} & \textbf{MC2} & \textbf{MC3} & \textbf{News} & \textbf{Wiki} & \textbf{Expert} & \textbf{StrQA} & \textbf{GSM8K}\\
\hline 
\rowcolor{gray!20}\multicolumn{9}{c}{\textbf{LLAMA3-8B-Instruct}} \\
Base  & 43.13	&61.26	&33.89	&\textbf{70.44}	&59.15	&63.78&	\textbf{72.67}&	75.78\\

Base + \textbf{PLI}  & \textbf{43.90}	& \textbf{61.63}	& \textbf{34.21}	& 70.11& 	\textbf{59.62}& 	\textbf{64.22}& 	72.41	& \textbf{76.29}\\

\hline

ITI & 43.39	&61.53	&33.94	&60.19	&47.22	&52.76&	68.17&69.20\\

ITI + \textbf{PLI} & \textbf{43.76}& \textbf{62.74}&	\textbf{35.21}&	\textbf{62.07}&	\textbf{49.37}&	\textbf{53.60}&	\textbf{68.30}&	\textbf{70.21}\\

\hline

SH2 & 43.30&	64.47&	36.23&	70.80&	59.69&	64.22& \textbf{72.38}&	76.92\\

SH2 + \textbf{PLI}& \textbf{45.62}&	\textbf{66.20}&	\textbf{38.57}&	\textbf{71.24}&	\textbf{60.08}&	\textbf{64.76}&	72.13&	\textbf{77.23}\\

\hline

DoLa & 42.96&65.76&35.71&70.10&59.37&64.06&72.22&76.30\\

DoLa + \textbf{PLI} & \textbf{45.60} &	\textbf{67.34} &	\textbf{37.62} & \textbf{70.53}&	\textbf{59.84} &\textbf{64.65}	& \textbf{72.61}&	\textbf{76.92}\\

\hline

ICD & 61.76	&\underline{\textbf{79.63}}	&58.90	&71.99	&60.55	&65.51&	72.45	&77.35\\

ICD + \textbf{PLI} & \underline{\textbf{63.35}}&79.22&	\underline{\textbf{59.47}}&	\underline{\textbf{72.42}}&	\underline{\textbf{61.10}}&	\underline{\textbf{65.87}}&	\underline{\textbf{72.85}}&	\underline{\textbf{77.92}}\\

\hline 

\rowcolor{gray!20}\multicolumn{9}{c}{\textbf{Mistral-7B-Instruct-v0.2}} \\

Base  & 55.26&72.08&44.33&74.50&\textbf{60.38}&65.51&\textbf{67.87}&43.09\\

Base + \textbf{PLI}&  \textbf{56.00}&\textbf{72.36}&	\textbf{45.71}&	\textbf{75.20}&	59.63&	\textbf{66.12}&	67.55&	\textbf{43.54}\\

\hline

ITI & 55.12	&72.30	&44.87	&68.76	&56.74	&59.82&	62.33	&\textbf{39.70}\\

ITI + \textbf{PLI} & \textbf{56.20}&	\textbf{72.82}&	\textbf{45.21}&	\textbf{69.30}&	\textbf{57.35}&	\textbf{61.47}&	\textbf{62.89}&	39.44\\

\hline

SH2 & 54.29	&75.32	&47.80	&74.93	&61.02	&67.40&	67.54	&43.37\\

SH2 + \textbf{PLI}& \textbf{55.43}&	\textbf{76.10}&	\textbf{49.75}&	\textbf{75.24}&	\textbf{61.40}&	\textbf{67.82}&	\textbf{67.93}&	\underline{\textbf{43.62}}\\

\hline

DoLa & 52.19&77.85&48.21&\textbf{73.82}&60.62&66.70&67.74&42.33\\

DoLa + \textbf{PLI} & \textbf{54.35}&\textbf{78.32}	&\textbf{48.70}	&73.25&	\textbf{61.02}	&\textbf{67.22}	&\textbf{68.01}	&\textbf{42.70}\\

\hline

ICD & 62.62	&79.83	&56.37	&75.76	&60.92	&\underline{\textbf{68.95}}&	68.17	&42.07\\

ICD + \textbf{PLI} & \underline{\textbf{65.56}}	&\underline{\textbf{80.49}}	&\underline{\textbf{58.97}}	&\underline{\textbf{76.20}}&	\underline{\textbf{61.73}}	&68.52	&\underline{\textbf{69.42}}	&\textbf{42.93}\\

\bottomrule

\end{tabular}
\end{center}
\caption{Experimental results on LLAMA3-8B-Instruct and Mistral-7B-Instruct-v0.2 across TruthfulQA, FACTOR, StrategyQA (StrQA), and GSM8K datasets. The \textbf{bolded} values indicate the better result in pairwise comparisons, while the \textbf{bolded} and \underline{\textbf{underlined}} values represent the best overall result for each benchmark. Our method achieves superior performance in most cases, demonstrating its effectiveness in hallucination mitigation and reasoning.}
\label{tab1}
\end{table*}

\subsection{Baselines}

To evaluate the effectiveness of our proposed PLI, we compare it against the following hallucination mitigation methods: \textbf{Base}: The original LLM without any modifications. \textbf{CD} \cite{li2023contrastive}: A classic CD method using two different-scale LLMs of the same type. Due to model size constraints, following \cite{zhang2023alleviating, zhang2024truthx}, we perform CD between 13B-Chat and 7B-Chat only on LLAMA2. \textbf{ITI} \cite{li2023inference}: A method that mitigates hallucinations by editing the activation of attention heads during inference. \textbf{DoLa} \cite{chuang2023dola}: A CD approach that contrasts early-exit layers with the final output layer to improve factual accuracy. \textbf{SH2} \cite{kai2024sh2}: A method that introduces low-confidence tokens at the inference stage to adjust output probabilities, encouraging the model to reassess its responses for improved truthfulness. \textbf{ICD} \cite{zhang2023alleviating}: A state-of-the-art CD method that reduces hallucination by contrasting the base model with a hallucinated version.



\subsection{Experiments Results}    \label{4.4}

Table \ref{tab1} present the performance of baseline methods and their PLI-enhanced variants, demonstrating that our proposed PLI significantly improves the performance of all three base models and existing hallucination mitigation techniques across multiple datasets. Additional results are provided in §\ref{A.1}.  These results highlight PLI's effectiveness in enhancing factuality in LLMs through the insertion of premature layers, calculated via mathematical interpolation. Further details on completion details and insertion position selection are provided in §\ref{A.2}.

Specifically, on TruthfulQA, PLI consistently improves performance across all methods. Notably, on LLAMA3-8B, it yields the largest gains for ICD (+1.59 points in MC1) and DoLa (+2.64 points in MC1). On FACTOR, PLI achieves consistent, albeit smaller, improvements for all base models, particularly in the News and Wiki subsets. This suggests that its impact is most pronounced in tasks requiring fine-grained factual reasoning. For reasoning and generation tasks on
StrategyQA and GSM8K, PLI also demonstrates notable gains, improving accuracy by up to 0.5 points on StrategyQA and an average of 0.4 points on GSM8K for LLAMA3-8B-Instruct and Mistral-7B-Instruct. This underscores PLI’s broader applicability beyond factuality enhancement. However, results for CD are mixed. While it performs well in some cases, it shows slight degradation on LLAMA2-7B-Chat ($-0.45$ points in MC1 and a similar decline on FACTOR). This is likely due to conflicts between CD’s contrastive mechanism and PLI’s interpolation-based approach. ITI, which modifies model activations, can be prone to perturbations. However, PLI assists it to achieve more improvement. Furthermore, when integrated with ICD, a strong CD method, PLI achieves the best overall performance in most cases. An extra analysis of statistical significance is conducted in Section §\ref{Statistical}.


\section{Analysis}
\subsection{Inference Consumption Testing}
Since our proposed method inserts premature layers, calculated via mathematical interpolation, into LLMs, it is important to assess its impact on inference efficiency. To evaluate this, we measure the inference time and GPU memory usage of LLMs before and after applying PLI, comparing these results with other hallucination mitigation methods.
For this analysis, we use LLAMA3-8B-Instruct as an example, with results shown in Table \ref{tab2}. Here, \textbf{PLI*n} denotes the number of inserted layers, and the integrations with DoLa and ICD remain consistent with the main experiment. We report the average inference time per sample of data and calculate the percentage of memory increase.

\textbf{Minimal impact on GPU memory usage:} PLI does not significantly affect memory consumption, as it only inserts specific layers during inference time, leaving model loading and tokenization largely unchanged. 

\textbf{Inference time remains efficient:} While PLI introduces slight overhead due to additional layers  and parameters, the impact is relatively small compared to its factuality improvements.

Compared with other methods, DoLa has slightly higher inference costs than the base model due to contrastive decoding between the final and early-exit layers. ICD incurs substantially higher computational costs, as it requires loading two models and computing their logits separately. 

\begin{table}[t] \footnotesize
\setlength\tabcolsep{0.7pt}
\renewcommand\arraystretch{1.1}
\begin{center}
\begin{tabular}{lcccc}
\toprule
\multirow{2}{*}{\textbf{Method}}&\multicolumn{3}{c}{\textbf{Datasets (s/sample)}}&\multirow{2}{*}{\textbf{Memory (Mib)}} \\
\cline{2-4}  
& \textbf{TruthfulQA} & \textbf{StrQA} & \textbf{GSM8K}  \\
\hline

Base & 0.042 &0.114 &0.032 & 17,143\\
Base + \textbf{PLI*1} & \textbf{0.046}&\textbf{0.118}&\textbf{0.033} &\textbf{17,430(+1.67\%)}\\
Base + \textbf{PLI*2} & \textbf{0.047} &\textbf{0.124}&\textbf{0.034} &\textbf{17,486(+2.00\%)}\\
Base + \textbf{PLI*3} & \textbf{0.049}&\textbf{0.127}&\textbf{0.035} &\textbf{17,673(+3.09\%)}\\
\hline \hline
Dola & 0.045&0.117&0.033&17,213\\
Dola + \textbf{PLI} & \textbf{0.048}	&\textbf{0.126}&	\textbf{0.034}&\textbf{17,463(+1.45\%)}\\
\hline \hline
ICD & 0.094&0.243&0.065 & 34,788\\
ICD + \textbf{PLI} & \textbf{0.098}&\textbf{0.254}&\textbf{0.069}& \textbf{35,023(+0.67\%)}\\

\bottomrule

\end{tabular}
\caption{Inference time and memory usage across different settings of three benchmarks based on LLAMA3-8B-Instruct.}
\label{tab2}
\end{center}
\end{table}

\begin{figure*}[t]
	\centering
\includegraphics[width=1\linewidth]{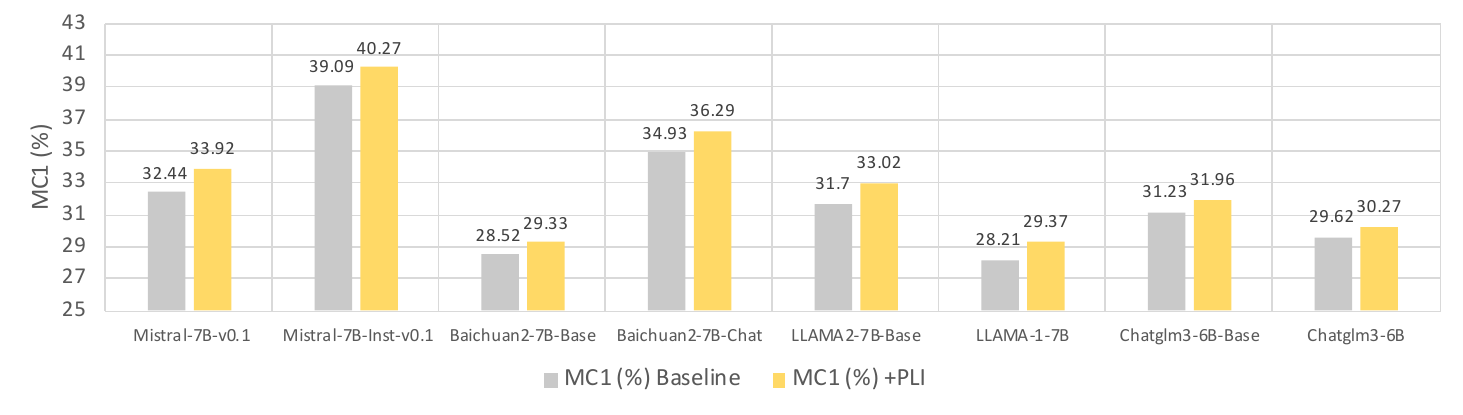}
	\caption{Experimental results on TruthfulQA (MC1 \%) across eight different base models.}
	\label{f3}
\end{figure*}

\subsection{Adaptability Testing on More Base Models}
To assess the adaptability of our method, we apply PLI to eight different LLMs and report the performance improvements on the TruthfulQA benchmark in Figure~\ref{f3}. Experimental results demonstrate that PLI consistently mitigates hallucinations and enhances factuality across various model architectures and scales, with minimal computational overhead due to its plug-and-play nature. Notably, PLI achieves a performance gain of $1.48$ points on Mistral-7B-v0.1 and $1.32$ points on LLAMA2-7B-Base. On average, PLI improves performance by $1.08$ points across the eight evaluated LLMs.

\subsection{Interpolation Techniques Comparison}
The proposed PLI method uses \textbf{Slerp} (spherical linear interpolation) as the interpolation strategy, as detailed in Section~\ref{3.1}. In this section, we compare the impact of different mathematical interpolation methods on the performance of PLI. The following interpolation methods are considered:

\textbf{Linear Interpolation (Lerp):} This method is computationally efficient but lacks the directional consistency needed in high-dimensional spaces. The parameters for interpolation are computed as:
\begin{equation}
\mathbf{w}_\mathrm{inter}=(1-\alpha)\mathbf{w}_l+\alpha \mathbf{w}_{l+1}
\end{equation}
where $\mathbf{w}_l$ and $\mathbf{w}_{l+1}$ are the parameters of the layers at positions $l$ and $l+1$, and $\alpha$ is the interpolation ratio.

\textbf{Bézier Curve Interpolation (BCerp):} This interpolation method adds curvature continuity by extending the linear interpolation to a quadratic Bézier curve with an additional control point $\mathbf{h}$, making the transitions smoother but more computationally expensive. The interpolated parameters are calculated as:
\begin{equation}
\mathbf{w}_\mathrm{inter}=(1-\alpha)^2\mathbf{w}_{l}+2\alpha(1-\alpha)\mathbf{h}+\alpha^2\mathbf{w}_{l+1}
\end{equation}
\begin{equation}
\mathbf{h}=(\mathbf{w}_{l}+\mathbf{w}_{l+1})/2
\end{equation}
where $\alpha$ is the interpolation ratio, and $\mathbf{h}$ is the control point for the curve. 

To compare these interpolation techniques, we conduct experiments on TruthfulQA using LLAMA3-8B and LLAMA2-7B-Chat, integrating the interpolation methods with ICD and a mixture of interpolation methods. The experimental results are shown in Table~\ref{tab3}.
\textbf{PLI} refers to the results obtained with the standard PLI approach (using \textbf{Slerp} as in the main experiments). \textbf{PLI (Mix)} integrates different interpolation techniques for various layers: \textbf{Slerp} is used in lower layers (e.g., the 8th layer), and \textbf{Lerp} is applied to higher layers (e.g., the 24th and 28th layers). The results show that all interpolation strategies help alleviate hallucination, outperforming ICD in most cases. The \textbf{PLI (Mix)} approach achieves promising performance, likely because the use of \textbf{Slerp} in the low layers preserves semantic consistency across the dimensional space, while \textbf{Lerp} in the higher layers efficiently maintains factual consistency with fewer disturbances.

\begin{table}[t] \footnotesize
\setlength\tabcolsep{10pt}
\renewcommand\arraystretch{1.1}
\begin{center}
\begin{tabular}{lccc}
\toprule
\multirow{2}{*}{\textbf{Method}}&\multicolumn{3}{c}{\textbf{TruthfulQA}} \\
\cline{2-4}  
& \textbf{MC1} & \textbf{MC2} & \textbf{MC3}  \\
\hline

\rowcolor{gray!20}\multicolumn{4}{c}{\textbf{LLAMA3-8B-Instruct}} \\
Base & 43.13&61.26&33.89\\

ICD & 61.76	&\textbf{79.63}	&58.90\\

ICD + \textbf{PLI}& 63.35 &79.22 &\textbf{59.47}\\

ICD + \textbf{PLI (Lerp)}& \textbf{64.37} &78.72 &58.22\\

ICD + \textbf{PLI (BCerp)}& 63.10 &78.13 &58.45\\

ICD + \textbf{PLI (Mix)}& 63.56 &78.27 &59.04\\

\hline
\rowcolor{gray!20}\multicolumn{4}{c}{\textbf{LLAMA2-7B-Chat}} \\
Base & 37.00 &	54.65 &27.82\\

ICD & 45.09	&69.10	&41.59\\

ICD + \textbf{PLI}& 46.69  &\textbf{70.70}&\textbf{43.52}\\

ICD + \textbf{PLI (Lerp)}& 46.81&69.70&42.53\\

ICD + \textbf{PLI (BCerp)}& 47.42&	69.72  &42.78\\

ICD + \textbf{PLI (Mix)}& \textbf{48.03}	&70.49	&43.05\\

\bottomrule

\end{tabular}
\caption{Experimental results on TruthfulQA with different interpolation techniques applied to LLAMA3-8B-Instruct and LLAMA2-7B-Chat.}
\label{tab3}
\end{center}
\end{table}

\subsection{PLI Mechanism Exploration}
This section explores the mechanism of the PLI method by analyzing how it interacts with the internal functioning of LLMs. Specifically, we investigate the relationship between hidden states output by consecutive layers and the effectiveness of PLI.
The experiment begins by defining two normal distributions based on the mean and standard deviation of the hidden states output by each layer and its subsequent layer. The Kullback-Leibler (KL) divergence is calculated between these distributions to gauge how they change under different interpolation schemes. Notably, the KL divergence between the second-to-last and last layers provides insights into the effectiveness of PLI.
Results on the TruthfulQA benchmark, shown in Table~\ref{tab4}, demonstrate that when PLI is effective, the KL divergence between the second-to-last and the last layer increases significantly. Conversely, when PLI is less effective, the KL divergence decreases. This suggests that PLI influences how the final layer of the model outputs tokens, with a higher divergence indicating that the final layer is better at outputting factual information. The final layer tends to restore tokens according to language priors, and PLI enhances its ability to output factually accurate words, increasing the KL divergence between the hidden states of the second-to-last and last layers.
This finding helps explain why the PLI method is effective in enhancing the factuality of LLMs: by adjusting the hidden states through premature layers, PLI helps the final layer output more accurate and factually consistent tokens.

\begin{table}[t] \footnotesize
\setlength\tabcolsep{3pt}
\renewcommand\arraystretch{1.1}
\begin{center}
\begin{tabular}{lcccc}
\toprule
\multirow{2}{*}{\textbf{Method}}&\multicolumn{3}{c}{\textbf{TruthfulQA}}&\multirow{2}{*}{\textbf{KL Div}} \\
\cline{2-4}  
& \textbf{MC1} & \textbf{MC2} & \textbf{MC3}  \\
\hline
\rowcolor{gray!20}\multicolumn{5}{c}{\textbf{LLAMA2-7B-Chat}} \\

Base & 37.00&54.65&27.82 & 153.6\\
Base + \textbf{PLI ($l_{i}=24,28$)} & \textbf{37.62}&\textbf{54.99}&\textbf{28.32} & \textbf{178.2(↑)}\\
Base + \textbf{PLI ($l_{i}=4,8$)} & 35.41&	54.23&	27.15 & 123.3(↓)\\
\hline 
\rowcolor{gray!20}\multicolumn{5}{c}{\textbf{Mistral-7B-Instruct-v0.2}} \\
Base & 55.26&72.08&44.33 & 132,096\\
Base + \textbf{PLI ($l_{i}=24,28$)} & \textbf{56.00}&\textbf{72.36}&\textbf{45.71} & \textbf{174,080(↑)}\\
Base + \textbf{PLI ($l_{i}=4,8$)} & 54.32 &71.21 &43.86 & 122,856(↓)\\

\bottomrule

\end{tabular}
\caption{Changes in the KL divergence of the hidden states distribution of the second-to-last and last layer of the model on the TruthfulQA under different PLI schemes.}
\label{tab4}
\end{center}
\end{table}

\begin{table}[t] \footnotesize
\setlength\tabcolsep{8.5pt}
\renewcommand\arraystretch{1.1}
\begin{center}
\begin{tabular}{lccc}
\toprule
\multirow{2}{*}{\textbf{Method}}&\multicolumn{3}{c}{\textbf{TruthfulQA}} \\
\cline{2-4}  
& \textbf{MC1} & \textbf{MC2} & \textbf{MC3}  \\
\hline

\rowcolor{gray!20}\multicolumn{4}{c}{\textbf{LLAMA3-8B-Instruct}} \\
Base (final layer) & 43.13&61.26&33.89\\
Base + \textbf{PLI} (final layer) & \textbf{43.90}&\textbf{61.63}&\textbf{34.21}\\
\hline \hline
Base (26th layer) & \textbf{24.63}	&50.79	&25.46\\
Base + \textbf{PLI} (\textbf{27th layer}) & 23.89&\textbf{50.85}&\textbf{25.50}\\
\hline \hline
Base (30th layer)&26.83&52.23&26.64\\
Base + \textbf{PLI} (\textbf{32nd layer}) & \textbf{27.94}	&\textbf{53.03}	&\textbf{27.31}\\

\hline
\rowcolor{gray!20}\multicolumn{4}{c}{\textbf{Mistral-7B-Instruct-v0.2}} \\
Base (final layer) & 55.26&72.08&44.33\\
Base + \textbf{PLI} (final layer) & \textbf{56.00}&\textbf{72.36}&\textbf{45.71}\\
\hline \hline
Base (26th layer) & \textbf{25.00}&50.12	&25.57\\
Base + \textbf{PLI} (\textbf{27th layer}) & 24.75	&\textbf{50.21}	&\textbf{25.98}\\
\hline \hline
Base (30th layer)& 25.36&51.28&26.66\\
Base + \textbf{PLI} (\textbf{32nd layer}) & \textbf{25.85}	&\textbf{51.41}	&\textbf{26.72}\\

\bottomrule

\end{tabular}
\caption{Experimental results of the final layer and early-exit layer of the models before and after PLI execution (both $l_{i}=24,28$) on TruthfulQA.}
\label{tab5}
\end{center}
\end{table}

Furthermore, we conduct an experiment, as shown in Table \ref{tab5}. Specifically, we evaluate the output of early-exit layers (26th and 30th selected, becoming the 27th and 32nd after PLI) and the final layer from LLMs before and after PLI execution  (with insertion locations $l_{i}=24,28$). The results show that, in addition to improving the factuality of the final layer, PLI also enhances the output of the premature layers within the model. This indicates that our method effectively optimizes the output representation of the middle layers by expanding the information processing flow through the insertion of premature layers, which helps alleviate hallucination in the final layer's output.

\subsection{Case Study}

To further validate the effect of our proposed method, we conduct a case study on the generated content before and after the execution of PLI based on LLAMA3-8B-Instruct, using StrategyQA, a benchmark focused on generation and logical reasoning. The case study is summarized in Table~\ref{tab6}. We observe that PLI significantly improves the factuality of the generated text by inserting premature layers into the LLMs. Additionally, PLI causes minimal interference, as evidenced by the high consistency between the two types of generated text. This demonstrates that PLI alleviates hallucination in the content without affecting the overall quality of the text generation.

\begin{table*}[t] \footnotesize
\setlength\tabcolsep{15pt}
\renewcommand\arraystretch{1.1}
\begin{center}
\begin{tabular}{p{6.5cm}p{6.5cm}}
\toprule
\textbf{LLAMA3-8B-Instruct} & \textbf{LLAMA3-8B-Instruct + PLI}\\
\hline
\multicolumn{2}{c}{\textbf{\textit{Question:}} \textit{Can an art dealer buy Boeing 737-800 with a Da Vinci painting?}} \\
\hline
The Boeing 737-800 is a commercial airliner. It costs tens of millions of dollars. A Da Vinci painting is a work of art. It could cost tens of millions of dollars. \textcolor{BrickRed}{However, it is unlikely that an art dealer would have the funds to buy a Boeing 737-800 with a Da Vinci painting.} So the answer is \textcolor{BrickRed}{no}. & The price of a Boeing 737-800 is around \textcolor{ForestGreen}{\$80 million}. The price of a Da Vinci painting can be tens of millions of dollars, but it would likely be less than \$80 million. Thus, \textcolor{ForestGreen}{an art dealer could potentially buy a Boeing 737-800 with a Da Vinci painting.} So the answer is \textcolor{ForestGreen}{yes}.   \\
\hline
\multicolumn{2}{c}{\textbf{\textit{Question:}} \textit{Would a clouded leopard encounter an awake pangolin?}} \\
\hline
Clouded leopards are found in Southeast Asia, while pangolins are found in Africa. These are two different continents. \textcolor{BrickRed}{Thus, it is unlikely that a clouded leopard and an awake pangolin would encounter each other.} So the answer is \textcolor{BrickRed}{no}. & Clouded leopards are found in Southeast Asia and China. Pangolins are found in Africa and \textcolor{ForestGreen}{Asia}. There is some overlap in their ranges, but they are not the same. \textcolor{ForestGreen}{Thus, it is possible for a clouded leopard and an awake pangolin to encounter each other}. So the answer is \textcolor{ForestGreen}{yes}.   \\
\hline
\multicolumn{2}{c}{\textbf{\textit{Question:}} \textit{Is Phobos (moon) name origin similar to Roman god Pavor?}} \\
\hline
Phobos is the name of a moon of Mars. Phobos is the Greek god of fear. Pavor is the Latin word for fear. The name Phobos comes from Greek mythology, not Latin. Thus, \textcolor{BrickRed}{the name origin is not similar.} So the answer is \textcolor{BrickRed}{no}. & Phobos is the Greek word for fear. Pavor is the Latin word for fear. \textcolor{ForestGreen}{Both words come from the same root.} Thus, the name origin of Phobos \textcolor{ForestGreen}{is similar to the Roman god Pavor}. So the answer is \textcolor{ForestGreen}{yes}.   \\

\bottomrule

\end{tabular}
\caption{Case study of PLI, showcasing the generation results of hallucination alleviation using PLI based on LLAMA3-8B-Instruct on StrategyQA. \textcolor{ForestGreen}{Green text} denotes factual content, while \textcolor{BrickRed}{red text} indicates hallucinated content.}
\label{tab6}
\end{center}
\end{table*}

\section{Conclusion}
In this paper, we introduce a novel method called PLI (\underline{P}remature \underline{L}ayers \underline{I}nterpolation), designed to enhance the factuality of LLMs. PLI is a training-free, plug-and-play intervention that optimizes the output representation of intermediate layers by inserting premature layers calculated through mathematical interpolation. This process expands the information flow within the model, thereby improving the factual accuracy of the final output and alleviating hallucinations. Our experiments on TruthfulQA, FACTOR, StrategyQA, and GSM8K datasets demonstrate that PLI outperforms other baseline methods in most cases. Additionally, we conduct a detailed analysis to explore the mechanism and effectiveness of the proposed approach.

\section{Limitations}
While PLI alleviates hallucinations by inserting premature layers constructed through mathematical interpolation, it introduces additional computational overhead, with the extent of this overhead varying across different models and tasks. Previous work \cite{geva2023dissecting} has pointed out that the high layers within models are tend to process and utilize factual information, while the information in the high layers is more redundant \cite{gromov2024unreasonable}, which may lead to our method having less impact on model performance in some cases. In the future, more explorations are needed for the internal mechanism of the model from the perspective of interpretability to further refine the PLI.

\section*{Acknowledgement}
This work was supported by Innovation Team Project of Guangdong Province of China (No. 2024KCXTD017), Shenzhen Science and Technology Foundation (No. JCYJ20240813145816022), National Key Research and Development Program of China (2024YFF0908200), National Natural Science Foundation of China (Grant No. 62376262) and the Natural Science Foundation of Guangdong Province of China (2024A1515030166, 2025B1515020032).

\bibliography{anthology,custom}

\clearpage

\appendix

\section{Appendix}

\subsection{The Experimental Results based on LLAMA2-7B-Chat} \label{A.1}

We present our experimental results based on LLAMA2-7B-Chat on the TruthfulQA, Factor, StrategyQA and GSM8K benchmarks, which are shown in Table \ref{tab9} and illustrated in §\ref{4.4}.

\begin{table*}[t] \small
\setlength\tabcolsep{11pt}
\renewcommand\arraystretch{1.2}
\begin{center}
\begin{tabular}{lcccccccc}

\toprule
\multirow{2}{*}{\textbf{Method}}&\multicolumn{3}{c}{\textbf{TruthfulQA}}&\multicolumn{3}{c}{\textbf{FACTOR}} &\multicolumn{2}{c}{\textbf{CoT}} \\
\cmidrule(r){2-4}  \cmidrule(r){5-7} \cmidrule(r){8-9}
& \textbf{MC1} & \textbf{MC2} & \textbf{MC3} & \textbf{News} & \textbf{Wiki} & \textbf{Expert} & \textbf{StrQA} & \textbf{GSM8K}\\
\hline 
\rowcolor{gray!20}\multicolumn{9}{c}{\textbf{LLAMA2-7B-Chat}} \\
Base  & 37.00&54.65	&27.82	&\textbf{64.71}	&56.61	&64.85&	63.67&	21.64\\

Base + \textbf{PLI}  & \textbf{37.62}	&\textbf{54.99}&	\textbf{28.32} &	64.45&	\textbf{56.90} &	\textbf{65.33}&	\textbf{64.03}&	\textbf{21.96}\\

\hline

ITI & 37.01	&54.66	&27.82	&53.28	&43.82	&51.69&	58.74	&\textbf{17.86}\\

ITI + \textbf{PLI} & \textbf{37.63}	&\textbf{55.48}	&\textbf{28.22}	&\textbf{54.52}&	\textbf{45.22}	&\textbf{53.47}	&\textbf{59.11}	&17.4\\

\hline

SH2 & 33.9	&57.07	&29.79	&65.31	&57.37	&\textbf{67.22}&	64.4 &\textbf{22.17}\\

SH2 + \textbf{PLI}& \textbf{35.49}&	\textbf{57.62}&	\textbf{30.44}&	\textbf{65.53}&	\textbf{57.64}&	66.89	&\textbf{64.81}	&22.09\\

\hline

CD & \textbf{28.15}	&54.87	&29.75	&\textbf{64.57}	&\underline{\textbf{58.47}}	&67.12	&58.42	&15.04\\

CD + \textbf{PLI}  & 27.70	&\textbf{55.92}	&\textbf{31.46}	&64.06&	58.03	&\textbf{67.33}	&\textbf{60.12}	&\textbf{16.30}\\

\hline

DoLa & 32.97&	60.84&	29.50&	64.32&	57.63&	67.30&	64.16&	22.07\\

DoLa + \textbf{PLI} & \textbf{34.79}&\textbf{62.10}	&\textbf{31.71}	&\textbf{64.75}&	\textbf{57.90}	&\textbf{68.43}	&\textbf{64.52}	&\textbf{22.49}\\

\hline

ICD & 45.09	&69.10	&41.59	&65.20	&56.57	&67.66&	64.37	&21.72\\

ICD + \textbf{PLI} & \underline{\textbf{46.69}}	&\underline{\textbf{70.70}}	&\underline{\textbf{43.52}}	&\underline{\textbf{65.73}}&	\textbf{56.98}	&\underline{\textbf{69.22}}	&\underline{\textbf{65.59}}	&\underline{\textbf{22.81}}\\

\bottomrule

\end{tabular}
\end{center}
\caption{Experimental results on LLAMA2-7B-Chat across TruthfulQA, FACTOR, StrategyQA (StrQA) and GSM8K datasets. To highlight, the \textbf{bolded} result indicates the better one in the pairwise comparison, while the \textbf{bolded} and \underline{\textbf{underlined}} result indicates the best for that benchmark. Our method outperforms other baselines in most cases across the four benchmarks.}
\label{tab9}
\end{table*}

\subsection{Completion Details} \label{A.2}
For our experiments, we use LLAMA3-8B-Instruct \cite{llama3modelcard}, Mistral-7B-Instruct-v0.2, and LLAMA2-7B-Chat \cite{touvron2023llama} as the base models.
Given that these models primarily consist of 32-layer transformers, we streamline the insertion of premature layers by following insights from previous work \cite{chuang2023dola}. Specifically, we select the following candidate insertion positions: $l_{i} = \{4, 8, 12, 16, 20, 24, 28\}$. These positions are chosen to align with the information distribution across layers in LLMs: low layers $\{4, 8\}$, middle layers $\{12, 16, 20\}$, and high layers $\{24, 28\}$, respectively. During experiments, we randomly insert 1–3 premature layers across different models and downstream tasks. The insertion ratio is dynamically computed using the scheduling formula from Section §\ref{3.3}, ensuring an extended information processing flow and optimized intermediate representations.
Since PLI is a plug-and-play method, we integrate it into the base models as well as various hallucination mitigation techniques to validate its effectiveness. All experiments are conducted on a single NVIDIA A800-80G GPU.

For the base models (i.e., greedy decoding) and contrastive decoding or representation editing methods within single models, we select insertion positions $l_{i} = \{24, 28\}$. For the ICD method, we use $l_{i} = \{8, 24, 28\}$ to emphasize the truthfulness of the models for a better contrast in results. All other experimental settings adhere to those used in previous works when comparing to other baseline methods.

While for the hyperparameters in the formula of our proposed Adaptive Interpolation Ratio Scheduling (as illustrated in \ref{3.3}). The constant $k$ is set to 4 by default to control the steepness of the sigmoid curve, while $c$ is set to 0.375 by default to determine the center offset of the function. Noting that the default settings of these two parameters are mainly suitable for LLMs with 32 layers. For models of larger sizes, $k$ and $c$ need to be adjusted to control the sigmoid curve and center offset, respectively. Nevertheless, it is convenient for mainly following the relationship that the inserted interpolation layer position and the interpolation ratio are positively correlated.

\subsection{Analysis of Layer Insertion Position}

This section explores the effect of our proposed PLI at different positions in the model. Taking the large language model of 32 layers of transformers as an example, we use LLAMA3-8B-Instruct to conduct experiments on the TruthfulQA benchmark with different interpolation combinations in $l_{i} = \{4, 8, 12, 16, 20, 24, 28\}$. The results are shown in Table \ref{tab12}.

\begin{table}[H] \footnotesize
\setlength\tabcolsep{7pt}
\renewcommand\arraystretch{1.1}
\begin{center}
\begin{tabular}{lccc}
\toprule
\multirow{2}{*}{\textbf{Method}}&\multicolumn{3}{c}{\textbf{TruthfulQA}} \\
\cline{2-4}  
& \textbf{MC1} & \textbf{MC2} & \textbf{MC3}  \\
\hline

\rowcolor{gray!20}\multicolumn{4}{c}{\textbf{LLAMA3-8B-Instruct}} \\
Base & 43.13&61.26&33.89\\

\textbf{PLI ($l_{i}=24,28$)} & \textbf{43.90}&\textbf{61.63}&\underline{34.21}\\

\textbf{PLI ($l_{i}=24$)} & 42.89&\underline{61.45}&\textbf{34.24}\\

\textbf{PLI ($l_{i}=12$)} & 42.60&61.07&33.95\\

\textbf{PLI ($l_{i}=12,16$)} & 43.13&61.23&33.79\\

\textbf{PLI ($l_{i}=12,16,28$)} & 42.76&60.95&33.57 \\

\textbf{PLI ($l_{i}=12,24,28$)} &\underline{43.42}&61.02&33.82 \\

\textbf{PLI ($l_{i}=12,16,24,28$)} &42.31&60.52&33.42 \\

\textbf{PLI ($l_{i}=12,16,20,24,28$)} &42.47&60.82&33.62 \\

\textbf{PLI ($l_{i}=4,12,16,24,28$)} &41.80&60.03&32.88 \\

\bottomrule

\end{tabular}
\caption{Experimental results on TruthfulQA with different settings of insertion position based on LLAMA3-8B-Instruct.}
\label{tab12}
\end{center}
\end{table}

Experimental results show that when the number of inserted interpolation layers is usually 2-3, PLI has a positive effect on hallucination alleviation, while the effect decreases when it is less or more. This may be because a single interpolation layer has little impact on large language models; yet more interpolation layers will inject more linear characteristics into LLMs, which will weaken the representation ability of the model. Meantime, it indicates that when the interpolation position is concentrated in the upper layer (such as $24$, $28$ layers), it results in better effectiveness, which may be because the upper layer of LLMs is closer to high-order semantic abstraction and factual information.

\subsection{Theoretical Interpretability of Inserted Premature Layers}

The PLI method we proposed mainly regards the LLMs from input to output as an information processing flow. From this perspective, there are two main reasons why our method is effective: (1) PLI extends the length of the information flow and the processing depth; (2) PLI maintains the continuity of the information flow and the knowledge manifold. Specifically, we are inspired by stable diffusion, which improves the quality of image processing generation by adjusting the sampling step. We migrate similar concepts to large language models and expand the number of model layers to extend the length of information flow transmission and promote the LLMs' capture and abstraction of information. On the other hand, the premature layers we inserted are calculated through mathematical interpolation (i.e., Slerp), which helps to maintain the directionality of vectors in high-dimensional space, and thus maintain the original manifold and continuity at the knowledge level. Our method needs to maintain a gbalance between the two properties to better alleviate hallucinations while maintaining the stability of the model architecture.

\subsection{Analysis of PLI with Models across Different Sizes}

We conducted experiments based on LLAMA2-13B-Chat (40 layers) and LLAMA2-70B-Chat (80 layers) on TruthfulQA dataset to demonstrate the adaptation of our proposed PLI on models across different sizes and layers, which are beyond 32 layers. The experimental results are shown in the following table \ref{tab14}.
\begin{table}[H] \footnotesize
\setlength\tabcolsep{12pt}
\renewcommand\arraystretch{1.1}
\begin{center}
\begin{tabular}{lccc}
\toprule
\multirow{2}{*}{\textbf{Method}}&\multicolumn{3}{c}{\textbf{TruthfulQA}} \\
\cline{2-4}  
& \textbf{MC1} & \textbf{MC2} & \textbf{MC3}  \\
\hline

\rowcolor{gray!20}\multicolumn{4}{c}{\textbf{LLAMA2-13B-Chat}} \\
Base & 37.62&54.60&28.12\\

Base + \textbf{PLI}& \textbf{38.80}	&\textbf{55.39}	&\textbf{28.92}\\

ICD & 48.47	&73.47	&46.04\\

ICD + \textbf{PLI}& \textbf{49.32}&\textbf{73.86}&\textbf{46.33}\\

\hline
\rowcolor{gray!20}\multicolumn{4}{c}{\textbf{LLAMA2-70B-Chat}} \\
Base & 38.79&58.99&30.59\\

Base + \textbf{PLI}& \textbf{39.52}&\textbf{59.40}&\textbf{31.06}\\

ICD & 53.24 &78.11 &49.14
\\

ICD + \textbf{PLI}& \textbf{54.07} &\textbf{79.60} &\textbf{49.54}\\

\bottomrule

\end{tabular}
\caption{Experimental results on TruthfulQA based on LLAMA2-13B-Chat and LLAMA2-70B-Chat.}
\label{tab14}
\end{center}
\end{table}

From the experimental results, we can see that our proposed method can perform well on larger size models. When the model itself contains more parameters, PLI need to insert more layers than the 7B model to optimize the performance of the model due to its original deeper information processing depth (for example, generally 5-6 layers are inserted on the 70B model). We insert 6 premature layers in our experiments based on LLAMA2-70B-Chat, mainly in layers 50-80, to achieve the results shown in our table.

\subsection{Experimental results on Open LLM Leaderboard}
In order to test PLI's impact on other capabilities with more benchmarks, we conduct experiments on Open LLM Leaderboard based on LLAMA3-8B-Instruct and Mistral-7B-Instruct-v0.2. The results are shown in the following table \ref{tab15}. Experimental results show that the proposed PLI can improve the performance of the original model in multiple test scenarios across different base models (bringing an average improvement of about 1-2\%), which further verifies the generalization of our method.

\begin{table*}[t] \footnotesize
\centering
\begin{tabular}{lccccccc}
\toprule
{\textbf{Method}} & {\textbf{Average}} & {\textbf{IFEval}} & {\textbf{BBH}} & {\textbf{MATH}} & {\textbf{CPQA}} & {\textbf{MUSR}} & {\textbf{MMLU-P}} \\
\midrule
LLAMA3-8B-Instruct         & 22.68 & 64.98 & 28.01 & 10.27 & 0.78 & 2.00 & 30.32 \\
LLAMA3-8B-Instruct + \textbf{PLI}     & \textbf{23.77} & \textbf{65.70} & \textbf{29.27} & \textbf{10.42}  & \textbf{1.96} & \textbf{2.75} & \textbf{32.54} \\
Mistral-7B-Instruct-v0.2   & 14.22 & 22.66 & 23.95 & 3.02  & 5.59 & 8.36 & 21.70 \\
Mistral-7B-Instruct-v0.2 + \textbf{PLI} & \textbf{15.01} & \textbf{23.42} & \textbf{24.79} & \textbf{3.33}  & \textbf{6.02} & \textbf{8.90} & \textbf{23.65} \\
\bottomrule
\end{tabular}
\caption{Experimental results on Open LLM Leaderboard based on LLAMA3-8B-Instruct and Mistral-7B-Instruct-v0.2.}
\label{tab15}
\end{table*}

\subsection{Analysis of Statistical Significance} \label{Statistical}

To further verify the stability of the PLI we proposed, we have selected several baselines, and ran 5 new rounds on TruthfulQA dataset based on LLAMA3-8B-Chat, calculated new means, then marked the numerical changes and p-test values, as shown in the following table \ref{tab16}.

\begin{table}[H] \footnotesize
\setlength\tabcolsep{3pt}
\renewcommand\arraystretch{1.1}
\begin{center}
\begin{tabular}{lccc}
\toprule
\multirow{2}{*}{\textbf{Method}}&\multicolumn{3}{c}{\textbf{TruthfulQA}} \\
\cline{2-4}  
& \textbf{MC1} & \textbf{MC2} & \textbf{MC3}  \\
\hline

\rowcolor{gray!20}\multicolumn{4}{c}{\textbf{LLAMA3-8B-Instruct}} \\
Base & 43.31\,($\pm$0.42) & 61.48\,($\pm$0.56) & 33.76\,($\pm$0.38)\\

Base + \textbf{PLI}& \textbf{44.12\,($\pm$0.47)} & \textbf{61.92\,($\pm$0.51)} & \textbf{34.43\,($\pm$0.45)}\\

p-value & \textit{0.018*}&\textit{0.043*}&\textit{0.022*} \\
\hline \hline

DoLa & 42.83\,($\pm$0.49) & 65.99\,($\pm$0.58) & 35.96\,($\pm$0.43)\\

DoLa + \textbf{PLI}& \textbf{45.83\,($\pm$0.53)} & \textbf{67.59\,($\pm$0.54)} & \textbf{37.48\,($\pm$0.47)}\\

p-value & \textit{0.0007**}&\textit{0.007**}&\textit{0.003**} \\
\hline \hline

ICD & 62.04\,($\pm$0.53) & \textbf{79.49\,($\pm$0.62)} & 58.72\,($\pm$0.48)\\

ICD + \textbf{PLI}& \textbf{63.65\,($\pm$0.52)} & 79.08\,($\pm$0.59) & \textbf{59.68\,($\pm$0.51)}\\

p-value & \textit{0.011*}&\textit{0.046*}&\textit{0.025*} \\

\bottomrule

\end{tabular}
\caption{Presentation of statistical significance on TruthfulQA based on LLAMA3-8B-Instruct.}
\label{tab16}
\end{center}
\end{table}

Noting that * indicates p-value < 0.05, ** indicates p-value < 0.01, and *** indicates p-value < 0.001 in the table, representing different statistical significance levels. It can be seen from the table that compared with the baselines, \textbf{improvements brought by PLI are basically significant}, which means that our method could stablely promote the baselines, verifying the robustness of our method.


\end{document}